\documentclass[sigconf]{acmart}

\AtBeginDocument{%
  \providecommand\BibTeX{{%
    \normalfont B\kern-0.5em{\scshape i\kern-0.25em b}\kern-0.8em\TeX}}}

\usepackage{subfigure}
\usepackage{multirow}

\usepackage{tikz}
\usepackage{algorithm}
\usepackage{algorithmic}
\usepackage{subfigure}
\usepackage{mathtools}

\usepackage{subfigure}
\usepackage{booktabs}
\usepackage{arydshln}
\usepackage{multirow}
\usepackage{bm}
\usepackage{color}
\usepackage{amsfonts}
\usepackage{mathbbol}
\usepackage{bm}
\usepackage[switch]{lineno}

\copyrightyear{2021} 
\acmYear{2021} 
\setcopyright{acmcopyright}\acmConference[KDD '21]{Proceedings of the 27th ACM SIGKDD Conference on Knowledge Discovery and Data Mining}{August 14--18, 2021}{Virtual Event, Singapore}
\acmBooktitle{Proceedings of the 27th ACM SIGKDD Conference on Knowledge Discovery and Data Mining (KDD '21), August 14--18, 2021, Virtual Event, Singapore}
\acmPrice{15.00}
\acmDOI{10.1145/3447548.3467451}
\acmISBN{978-1-4503-8332-5/21/08}

\begin{document}
\fancyhead{}

\title{Learning How to Propagate Messages in Graph Neural Networks}


\author{Teng Xiao$^{\S\ast}$, Zhengyu Chen$^{{\dagger}\S}$, Donglin Wang$^{\S\ddagger}$, and Suhang Wang$^{\ast}$
}
\thanks{$^{\ddagger}$Corresponding author.}
\affiliation{
   \institution{$^{\S}$Machine Intelligence Lab (MiLAB), AI Division, School of Engineering, Westlake University\\
    $^{\dagger}$College of Computer Science \& Technology, Zhejiang University  $^\ast$The Pennsylvania State University}
  \city{}
 \country{}
}
\email{tvx5054@psu.edu, {chenzhengyu,wangdonglin}@westlake.edu.cn, szw494@psu.edu}

\def\authors{Teng Xiao, Zhengyu Chen, Donglin Wang, and Suhang Wang}

\begin{abstract}
This paper studies the problem of learning  message propagation strategies for  graph neural networks (GNNs). One of the challenges for graph neural networks is that of defining the propagation strategy. For instance, the choices of propagation steps are often specialized to a single graph and are not personalized to different nodes.  To compensate for this,  in this paper, we present learning to propagate, a general learning framework that not only learns the GNN parameters for prediction but more importantly, can explicitly learn the interpretable and personalized  propagate strategies for different nodes and various types of graphs.
We introduce the optimal propagation steps as latent variables to help find the maximum-likelihood estimation of the GNN parameters in a variational Expectation-Maximization (VEM) framework.    
Extensive experiments on various types of graph benchmarks demonstrate that our proposed framework can significantly achieve better performance compared with the state-of-the-art methods, and can effectively learn personalized and  interpretable propagate strategies of messages in GNNs. Code is available at \url{https://github.com/tengxiao1/L2P}.
\end{abstract}

\ccsdesc[500]{Computing methodologies~Machine learning}

\keywords{Graph Neural Networks; Graph Representation Learning}


\maketitle

\section{Introduction}
\label{sec:intro}
Graphs are ubiquitous in the real world, such as  social networks, knowledge graphs, and molecular structures.  Recently, Graph Neural Networks (GNNs) have achieved state-of-the-art performance across various tasks on graphs, such as semi-supervised node classification~\cite{DBLP:conf/iclr/KipfW17,DBLP:conf/iclr/VelickovicCCRLB18,zhao2021graphsmote} and link prediction~\cite{hamilton2017inductive}.  Typically, GNNs exploit message propagation strategies to learn expressive node representations by  propagating and aggregating the messages between neighboring nodes. Various message propagation layers have been proposed, including graph convolutional layers (GCN)~\cite{DBLP:conf/iclr/KipfW17}, graph attention layers (GAT)~\cite{DBLP:conf/iclr/VelickovicCCRLB18}, and many others~\cite{DBLP:conf/iclr/KlicperaBG19,chen2020simple,xu2018representation,wu2019simplifying,hamilton2017inductive,dai2021say}.  Recent studies~\cite{li2018deeper,DBLP:conf/iclr/KipfW17,chen2020measuring} show that GNNs suffer from the over-smoothing issue (the representations of nodes are inclined to converge to a certain value, making the model performance degrade significantly by stacking too many propagation layers).

\begin{figure}
\setlength{\abovecaptionskip}{-0.2cm}
\setlength{\belowcaptionskip}{-0.4cm}
\centering
  \subfigure{
    \includegraphics[width=0.34\textwidth]{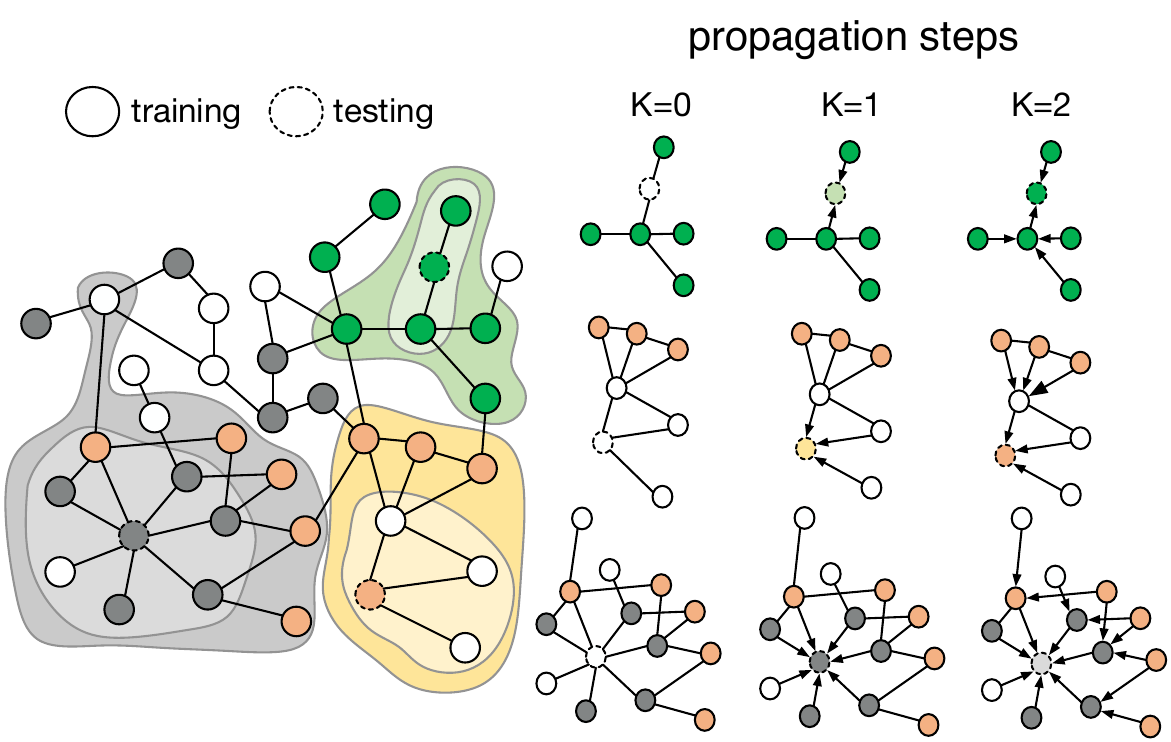}}  
\caption{ An illustration of the needing for personalized propagation. Color (\textcolor[RGB]{0,176,86}{green}, \textcolor[RGB]{130,133,133}{gray}, and \textcolor[RGB]{250,138,47}{yellow}) denotes the class of the node. White nodes denote the unlabeled nodes.}\label{fig:intro}
\end{figure}

To tackle the over-smoothing issue, many efforts have been taken~\cite{DBLP:conf/iclr/KipfW17,li2019deepgcns,xu2018representation,chen2018fastgcn,zhao2019pairnorm}. For example, several works~\cite{DBLP:conf/iclr/KipfW17,li2019deepgcns,xu2018representation} try to add residual or dense connections in the message propagation layer to  preserve the locality. Although the convergence speed of over-smoothing is retarded, most of these methods do not really outperform  2-layer models such as GCN or GAT. A crucial question remains to be addressed in order to make GNN a success: \textit{Do we really need a very deep GNN?  Furthermore, can we automatically  choose propagation steps to specific nodes and graphs?}

In this paper, we provide answers to both questions. Our key insight is that different nodes and various types of graphs may need different propagation steps to accurately predict node labels. As suggested  by~\cite{xu2018representation,DBLP:conf/cikm/TangYSWTAMW20,chen2020simple}, low-degree nodes only have a small number of neighbors, which receive very limited information from neighborhoods and deep GNNs may  perform well on those nodes. In contrast, nodes with higher degrees are more likely to suffer from over-smoothing.  Figure~\ref{fig:intro} illustrates the key motivation of this paper. Intuitively, the propagation step  smoothes the features locally along the edges of the graph and ultimately encourages similar predictions among locally connected nodes. We can obviously observe that the optimal propagation step for the test green and yellow class nodes is two, however, the optimal step for the black class node is one since more steps will bring the noise to the representation of it. Thus, it is a natural idea that different propagation patterns tend to work better for different nodes. For different types of graphs,  the number of propagation steps is also different. For example,  for those heterophily graphs~\cite{newman2002assortative} (where connected nodes may have different class labels and dissimilar features),  message propagation steps may hurt the node similarity, and stacking deep layers  cannot achieve better performance compared with homophily graphs~\cite{jin2020node, chen2020simple}.  Since there is no strategy to learn how  to propagate the message, existing GNNs need a hand-crafted layer number depending on different types of nodes and graphs. This requires  expert domain knowledge and careful parameter tuning and will be sub-optimal. However, whether it is possible to learn personalized  strategies while optimizing  GNNs remains an open problem.

Motivated by the discussion above, in this paper,  we investigate whether one can automatically learn personalized propagation strategies 
to help GNNs learn interpretable prediction and improve generalization. In essence, we are faced with several challenges: (i) The graph data is very  complex, thus building  hand-crafted and heuristic rules for propagation steps for each node  tends to be infeasible when we know little knowledge underlying graph or the node  distributions are too complicated. (ii)  In practice, there is no way to directly access the optimal strategy of propagation. The lack of supervision  about how  to propagate obstructs models from modeling the distribution of propagation for each node. (iii) GNNs are also prone to be over-fitting~\cite{DBLP:conf/iclr/RongHXH20}, where the GNNs fit the training data very well but generalizes poorly to the testing data. It will suffer from the over-fitting issue more severely when we utilize an addition parameterized  model for each node to learn how to propagate given limited labeled data in the real world.

To address the challenges mentioned above,  we propose a simple yet effective  framework called \text{learning to propagate} (L2P) to simultaneously learn the optimal propagation strategy and GNN parameters to achieve personalized and adaptive propagation. Our framework requires no heuristics and is generalizable to various types of nodes and graphs. Since there is no supervision, we  adopt the principle of the  probabilistic generative model and introduce the optimal propagation steps  as latent variables to help find the maximum-likelihood estimation of  GNN parameters in a variational Expectation-Maximization (VEM)  framework. To further alleviate the over-fitting, we introduce an efficient bi-level optimization algorithm. The bi-level optimization closely matches the definition of generalization  since validation data can provide accurate estimation of the generalization. The main contributions of this work are: \\
$\bullet$ We study a new problem of learning propagation strategies for GNNs. To address this problem, we propose a general L2P framework which can learn personalized and interpretable propagation strategies, and achieve better performance simultaneously.  \\
$\bullet$ We propose an effective stochastic algorithm based on variational inference and bi-level optimization for the L2P framework, which enables simultaneously learning the optimal propagation strategies and GNN parameters, and avoiding the over-fitting issue.  \\
$\bullet$ We conduct  experiments on homophily and heterophily graphs and the results demonstrate the effectiveness of our  framework. 

\section{Related Work}
\subsection{Graph Neural Networks}
GNNs have achieved great success in modeling graph-structured data. Generally, GNNs can be categorized into two categories, i.e., spectral-based and spatial-based. Spectral-based GNNs define graph convolution based on spectral graph theory~\cite{bruna2014spectral,defferrard2016convolutional,wu2019simplifying}. GCN~\cite{DBLP:conf/iclr/KipfW17} further simplifies graph convolutions by stacking layers of first-order Chebyshev polynomial filters together with some approximations. Spatial-based methods  directly define updating rules in the spatial space. For instance, GAT~\cite{DBLP:conf/iclr/VelickovicCCRLB18} introduces the self-attention strategy into aggregation to assign different importance scores of neighborhoods.   We refer  interested readers to the recent survey~\cite{wu2020comprehensive} for more variants of GNN architectures.
Despite the success of variants GNNs, the majority of existing GNNs aggregate neighbors’ information for representation learning, which are shown to suffer from the over-smoothing~\cite{oono2019graph,li2018deeper} issue when many propagation layers are stacked, the representations of all nodes become the same.

To tackle the over-smoothing issue, some works~\cite{DBLP:conf/iclr/KipfW17,li2019deepgcns} try to add residual or dense connections~\cite{xu2018representation} in propagation steps for preserving  the locality of the node representations. Other works~\cite{chen2018fastgcn,DBLP:conf/iclr/RongHXH20} augment the graph by randomly removing  a certain number of edges or nodes to prevent the over-smoothing issue. Recently, GCNII~\cite{chen2020simple} introduces initial residual and identity mapping techniques for GCN and  achieves promising performance.  Since the feature propagation  and transformation steps are commonly coupled with each other in standard GNNs, several works~\cite{DBLP:conf/iclr/KlicperaBG19, liu2020towards} separate this into two steps to reduce the risk of over-smoothing. We differ from these methods as (1) instead of focus  on alleviating over-smoothing, we argue that different nodes and graphs  may need a different number of propagation layers, and propose a framework of learning   propagation strategies that  generalizable to various types of graphs and backbones, and (2) we propose the bilevel optimization to utilize validation error to guide learning propagation strategy for improving the generalization ability of graph neural networks.

\subsection{The Bi-level Optimization}
Bi-level optimization~\cite{maclaurin2015gradient}, which performs upper-level learning subject to the optimality of lower-level learning, has  been applied to different tasks such as few-shot learning~\cite{finn2017model,chen2021multi,Chenpa21}, searching architectures~\cite{liu2018darts}, and reinforcement learning~\cite{zheng2018learning}. For the graph domain, \citeauthor{franceschi2019learning} propose a bi-level optimization objective  to learn the  structures of graphs. Some works~\cite{zhou2019auto,gao2019graphnas} optimize a bi-level objective via reinforcement learning to search the architectures of GNNs. Moreover, Meta-attack~\cite{zugner2018adversarial} adopts  the principle of meta-learning to conduct the poisoning attack on the graphs by optimizing a bi-level objective. Recently, \citeauthor{Hwang2020} propose SELAR~\cite{Hwang2020} which 
learns the weighting function for self-supervised tasks to help the primary task on the graph with a bi-level objective.  To conduct the few-shot learning in graphs, the work~\cite{liu2020towards}, inspired by MAML~\cite{finn2017model}, try to obtain a parameter
initialization that can adapt to unseen tasks quickly, using  gradients information from the bi-level optimization. By contrast, in this paper,  our main
concern is the generalization, and we propose a bilevel programming with variational inference to develop
a framework for learning  propagation strategies, while avoiding the over-fitting issues.

\section{Preliminaries}
\subsection{Notations and Problem Definition}
\label{sub:NP}
Let $G=(\mathcal{V}, \mathcal{E})$ denote a graph, where $\mathcal{V}$ is a set of $|\mathcal{V}|=N$ nodes and $\mathcal{E} \subseteq \mathcal{V} \times \mathcal{V}$ is a set of $|\mathcal{E}|$ edges between nodes. $\mathbf{A} \in\{0,1\}^{N \times N}$ is the adjacency matrix of $G$. The $(i,j)$-th element $\mathbf{A}_{i j}=1$ if there exists an edge between node $v_{i}$ and $v_{j},$ otherwise $\mathbf{A}_{i j}=0$. Furthermore, we use $\mathbf{X}=\left[\mathbf{x}_{1}, \mathbf{x}_{2}, \ldots, \mathbf{x}_{N}\right] \in \mathbb{R}^{N \times d}$ to denote the features of nodes, where $\mathbf{x}_n$ is the $d$-dimensional feature vector of node $v_n$. Following the common semi-supervised  node classification setting~\cite{DBLP:conf/iclr/KipfW17,DBLP:conf/iclr/VelickovicCCRLB18}, only a small portion of nodes $\mathcal{V}_{o}=\left\{v_{1}, v_{2}, \ldots, v_{o}\right\}$ are associated with observed  labels $\mathcal{Y}^{o}=\left\{y_{1}, y_{2}, \ldots, y_{o}\right\}$, where $y_{n}$ denotes the label of $v_{n}$. $\mathcal{V}_{u} = \mathcal{V}\backslash \mathcal{V}_o$ is the set of unlabeled nodes. Given the adjacency matrix $\mathbf{A}$,  features  $\mathbf{X}$ and  the observed labels $\mathcal{Y}^{o}$, the task of  node classification is to learn a function $f_{{\theta}}$ which can accurately predict the  labels $\mathcal{Y}^{u}$ of unlabeled nodes $\mathcal{V}_{u}$.

\subsection{Message Propagation}
\label{sec:MP}
Generally, GNNs adopt the  message propagation process, which iteratively aggregates the neighborhood information. Formally, the propagation process of the $k$-th layer in GNN is two steps:
\begin{linenomath}
\small
\begin{align}
&\mathbf{m}_{k,n}=\text{AGGREGATE }^{}\left(\left\{\mathbf{h}_{k-1,u}^{}: u \in \mathcal{N}(n)\right\}\right) \\
&\mathbf{h}_{k,n}=\text{UPDATE}^{}\left(\mathbf{h}_{k-1,n}^{},\mathbf{m}_{k,n}^{},\mathbf{h}_{0,n}^{}\right)
\end{align}
\end{linenomath}
where  $\mathcal{N}_{n}$ is the set of neighbors of node $v_{n}$, AGGREGATE is a permutation invariant function. After $K$ message-passing layers, the final node embeddings $\mathbf{H}_{K}$ are used to perform a given task. In general, most  state-of-the-art GNN backbones~\cite{DBLP:conf/iclr/KipfW17,DBLP:conf/iclr/VelickovicCCRLB18,DBLP:conf/iclr/KlicperaBG19,chen2020simple} follow this message propagation form with  different AGGREGATE functions, UPDATE function, or initial feature $\mathbf{h}_{0, n}^{}$. For instance, APPNP~\cite{DBLP:conf/iclr/KlicperaBG19} and GCNII~\cite{chen2020simple} add the initial feature $\mathbf{h}_{0,n}^{}=\text{MLP}(\mathbf{x}_{n}; \theta)$ to each layer in the UPDATE function. In general, GNN consists of several message propagation layers. We abstract the message propagation  with $K$ layers as one parameterized function $GNN(\mathbf{X},\mathbf{A},K)$.

\section{Learning to Propagate}
In this section, we introduce the $\text{Learning to Propagate}$ (L2P) framework, which can perform personalized message propagation  for better node representation learning and a more interpretable prediction process.  The key idea is to introduce a discrete latent variable $t_n$ for each node $v_n$, which denotes the personalized optimal propagation step of $v_n$. How to learn $t_n$ is challenging given no explicit supervision on the optimal propagation step of $v_n$. To address the challenge, we propose a generative model for modeling the 
 joint distribution of node labels and propagation steps conditioned on node attributes and graphs, i.e., $p(y_n,t_n|\mathbf{X},\mathbf{A})$, and formulate the \text{Learning to Propagate} framework as a variational objective, where the goal is to find the parameters of GNNs and the optimal propagation distribution, by iteratively approximating and maximizing the log-likelihood function. 
To alleviate the over-fitting issue, we further frame the variational process as a bi-level optimization problem, and  optimize the variational parameters of
learning the propagation strategies in an outer loop to maximize
generalization performance of GNNs trained based on the learned strategies. 
\subsection{The  Generative Process}
Generally, we can consider the designing process of graph neural networks as follows: we first choose the number of propagation layers $K$ for all nodes and the type of the aggregation function parameterized by $\theta$. Then for each training label ${y}_{n}$ of node $n$, we typically conduct the Maximum Likelihood Estimation (MLE) of the marginal log-likelihood over the observed labels as:
\begin{linenomath}
\small
\begin{align}
\max _{{\theta}} \mathcal{L}_{}\left(\theta; \mathbf{A}, \mathbf{X}, \mathcal{Y}^{o}\right)=
  \text{$\sum\nolimits_{y_{n} \in \mathcal{Y}^{o}}$} \log p_{{\theta}}\left(y_n| {GNN}\left(\mathbf{X}, \mathbf{A},K\right) \right), \label{Eq:GC}
\end{align}
\end{linenomath}where $p_{{\theta}}\left(y_n| {GNN}(\mathbf{X}, \mathbf{A}, K)\right)=p\left(y_{n} | \mathbf{H}_{K}\right)$ is the predicted probability of node $v_n$ having label $y_n$ using $\mathbf{H}_{K,n}$. $\mathbf{H}_{K,n}$ is the node representation of $v_n$ after stacking $K$ propagation steps (see \S~\ref{sec:MP}). Generally, a softmax is applied on $\mathbf{H}_{K,n}$ for predicting label $y_{n}$.

Although the message propagation strategy above has achieved promising performance, it has two drawbacks: (i) The above strategy treats each node equally, i.e., each node stacks $K$-layers; while in practice, different nodes may need different propagation steps/layers. Simply using a one-for-all strategy could potentially lead to sub-optimal decision boundaries and is less interpretable, and (ii) Different datasets/graphs may also have different optimal propagation steps. Existing GNNs require a hand-crafted  number of propagation steps, which requires expert domain knowledge, careful parameter tuning, and is time-consuming. Thus, it would be desirable if we could learn the personalized and adaptive propagation strategy which is applicable  to  various types of graphs and GNN backbones.

Based on the  motivation above, we propose to learn a personalized propagate distribution from the given labeled nodes and utilize the learned propagate distribution at test time, such that each test node would automatically find the optimal propagate step to explicitly improve the performance. 
A natural idea of learning optimal propagate distribution is supervised learning. However, there is no direct supervision of the optimal propagate strategy for each node.
To solve this challenge, we treat the optimal propagation layer of each node as a discrete latent variable and adopt the principle of the probabilistic generative model, which has shown to be effective in estimating the underlying data distribution~\cite{xiao2019hierarchical,chen2021deep,mohamed2015variational,xiao2021general}. 

Specifically, for each node $v_n$,
we introduce a latent discrete variable $t_{n}\in\{0, 1,2,\cdots, K\}$ to denote its optimal propagation step, where $K$ is the predefined maximum step. Note that $t_n$ can be $0$, which corresponds to use  non-aggregated features for prediction. We allow $t_n$ to be $0$, because for some nodes in  heterophily graphs, the neighborhood information is noisy, aggregating the neighborhood information may result in worse performance~\cite{pei2019geom,jin2020node}. $t_n$ is node-wise because the optimal propagation step for each node may vary largely from one node to another. With the latent variable $\{t_n\}_{n=1}^{|\mathcal{V}|}$, we propose the following generative model with modeling the joint distribution of each observed label $y_{n}$ and latent $t_n$:
\begin{linenomath}
\small
\begin{align}
p_{{\theta}}(y_{n},t_{n} | \mathbf{X},\mathbf{A})=p_{{\theta}}(y_{n} |  {GNN}(\mathbf{X},\mathbf{A}, t_{n}))p(t_{n}), \label{Eq:1}
\end{align}
\end{linenomath}
where $p(t_{n})$ is the prior of propagation variable and $\theta$ is the parameter shared by all nodes. $p_{{\theta}}(\mathbf{y}_{n} |  {GNN}(\mathbf{X},\mathbf{A},t_{n}))$ represents the label prediction probability using $v_n$'s representation from the $t_{n}$-th layer, i.e., $\mathbf{H}_{t_n,n}$.
Since we do not have any  prior of how to propagate, $p(t_{n})=\frac{1}{K+1}$ is defined as uniform distribution on all layers of all nodes in this paper. We can also use an alternative prior with lower probability on the deeper layers, if we want to encourage shallower GNNs.
Given the  generative model in Eq.~(\ref{Eq:1}) and from the Bayesian perspective, what we are interested in are two folds: \\
(1)  Learning the parameter $\theta$ of the GNN by maximizing the follow likelihood which  helps make label prediction in the testing phase:
\begin{linenomath}
\small
\begin{align}
\log p_{{\theta}}(y_{n} | \mathbf{X},\mathbf{A})= \log \text{$\sum\nolimits_{t_n=0}^{K}$} p_{{\theta}}(y_{n} |  {GNN}(\mathbf{X},\mathbf{A},t_{n}))p(t_{n}). \label{marginal log-likelihood}
\end{align}
\end{linenomath}
(2) Inferring the following posterior $p(t_n|\mathbf{X},\mathbf{A},y_n)$ of the latent variable $t$, which is related to the optimal propagation distribution.
\begin{linenomath}
\small
\begin{align}
p(t_n=k | \mathbf{X},\mathbf{A}, y_n)=\frac{p_{\theta}(y_{n} | GNN( \mathbf{X},\mathbf{A}, k))}{\sum_{k'=0}^{K} p_{\theta}(y_n | GNN(\mathbf{X},\mathbf{A}, k'))}. \label{Eq:non-par}
\end{align}
\end{linenomath}
Intuitively, this posterior can be  understood as we choose the propagation step $t_n$ of node $v_n$ according to the largest likelihood (i.e., the smallest loss) in the defined propagation steps.

However, there are several challenges to solve these two problems. For learning, we cannot directly learn the parameter $\theta$, since it involves marginalizing the latent variable, which is generally time-consuming and intractable~\cite{kingma2013auto}. In terms of the inference, since we do not have labels for test nodes, i.e, $y_n$ for $v_n \in \mathcal{V}_u$, the non-parametric true posterior in Eq.~(\ref{Eq:non-par}), which involves evaluating the likelihood $p_{\theta}(y_{n} | GNN( \mathbf{X},\mathbf{A}, k))$ of test nodes, is not applicable. To solve the challenges in the learning and inference, we adopt the variational inference principle~\cite{kingma2013auto,xiao2019bayesian}, and instead consider the following  lower bound of the marginal log-likelihood in Eq.~(\ref{marginal log-likelihood}) which  gives rise to our following formal variational  objective:
\begin{linenomath}
\small
\begin{align}
 \mathcal{L}({\theta}, q)= \mathbb{E}_{q_{}(t_n)}\left[\log p_{\theta}\left(\mathbf{y}_{n} | \mathbf{X}, \mathbf{A}, t_{n}\right)\right]-\text{KL}(q_{}(t_n) ||p(t_n)),\label{Eq:ELBO}
\end{align}
\end{linenomath}
where the derivations are given in Appendix~\ref{app:ELBO} and $q_{}(t_n)$ is the introduced variational distribution. Maximizing the  ELBO $\mathcal{L}({\theta}, q)$  is equivalent to (i) maximize Eq.~(\ref{marginal log-likelihood}) and to (ii) make the variational distributions $q_{}(t_n)$  of each node  be  close to its intractable true posteriors $p(t_n|\mathbf{X},\mathbf{A},y_n)$. Note that the ELBO   holds for any type of variational distribution $q_{}(t_n)$. We defer discussion of the learning and  inference process until the next section. Here, we first introduce two ways  to show how to exactly parameterize the variational distribution $q_{}(t_n)$, resulting in two instances of our L2P framework.
\subsection{Learning to Select}
In the variational inference principle, we can introduce a variational distribution $q_{\phi}(t_n| \boldsymbol{\nu}_{n})$ parameterized  by $\boldsymbol{\nu}_{n} \in \mathbb{R}^{K}$. However, we  cannot fit each  $q_{\phi}(t_n| \boldsymbol{\nu}_{n})$ individually by solving  $N \cdot K$ parameters,  which increases the over-fitting risk given the limited labels in the graphs. Thus, we consider the  amortization inference~\cite{kingma2013auto} which avoids the optimization of the parameter $\boldsymbol{\nu}_{n}$ for each local variational distribution $q_{\phi}(t_n| \boldsymbol{\nu}_{n})$. Instead, it fits a shared neural network to calculate each local parameter $\boldsymbol{\nu}_{n}$. Since the latent variable $t_{n}$ is a discrete multinomial variable, the simplest and most naive way to represent categorical variable is the softmax function. Thus, we pass the features of nodes through a softmax function to parameterize the categorical  propagation   distribution as:
\begin{linenomath}
\small
\begin{align}
q_{\phi}(t_n=k| \mathbf{X},\mathbf{A})=\frac{\exp (\mathbf{w}_{k}^{\top}\mathbf{H}_{k,n})}{ \sum_{k'=0}^{K}\exp (\mathbf{w}_{k'}^{\top}\mathbf{H}_{k',n})},
\end{align}
\end{linenomath}
where $\mathbf{w}_{k}$ represents the trainable linear transformation for the $k$-th layer. $\mathbf{H}_{k,n}$ is the representation of node $n$ at the $k$-th layer and $\phi$ represents the set of parameters. The main insight behind this amortization is to reuse the propagation representation of each layer, leveraging
the accumulated knowledge of representation  to quickly infer  propagation distribution. With amortization, we reduce the number of parameters to $(K+1)\cdot D$, where $K$ is the predefined  maximum propagation step and $D$ is the dimension of the representation of nodes. Since this formulation directly models the selection probability overall propagation steps of each node, we refer to this method as \textit{Learning to Select} (\text{L2S}). Figure~\ref{fig:framework}(b) gives an illustration of \text{L2S}.  We adopt the node representation of $v_n$ in each layer to calculate $q_{\phi}(t_n=k| \mathbf{X},\mathbf{A})$, which makes it able to personally and adaptively decide which propagation layer is best for $v_n$. It also allows each graph to learn its own form of propagation with its  own decay form from the validation signal (see \S~\ref{Main:Bilevel} for details).
\subsection{Learning to Quit}

\begin{figure}
\setlength{\abovecaptionskip}{-0.5cm}
\setlength{\belowcaptionskip}{-0.5cm}
\centering
  \subfigure{
    \includegraphics[width=0.32\textwidth]{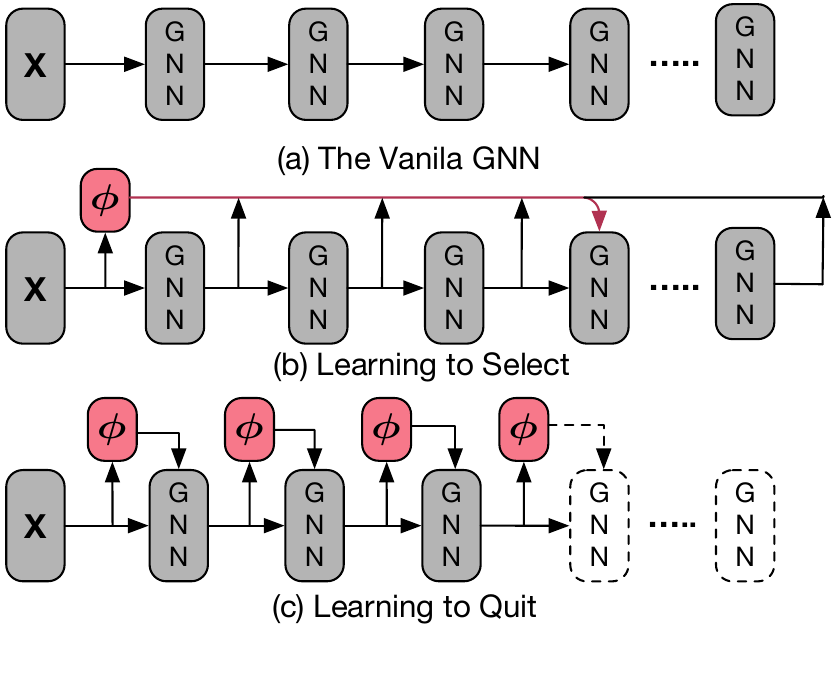}}
\caption{Illustrations of our L2P framework.  (a) The vanilla GNN architecture. (b) L2S predicts the selection probability over all propagation steps for each node. (c) L2Q forces each node to personally quit  its propagation process.}\label{fig:framework}
\end{figure}

\label{Sec:L2Q}
Instead of directly modeling the selection probability over every propagation step, we can model the probability of exiting the  propagation process and transform the modeling of multinomial probability parameters into the modeling of the logits of binomial probability parameters. More specifically, we consider modeling the quit probability at each propagation layer for each node $n$ as follows:
\begin{linenomath}
\small
\begin{align}
\alpha_{k,n}=\frac{1}{1+\exp({-\mathbf{w}_{k}^{\top}\mathbf{H}_{k,n})}},
\end{align}
\end{linenomath}
where $\alpha_{k,n}$ denotes the probability that node $v_n$ quits propagating at the $k$-th layer. The challenge is how to transfer the logits $\alpha_{k,n}$ to the multinomial variational distribution $q_{\phi}(t_n| \mathbf{X},\mathbf{A})$. In this paper, we consider the stick breaking (a non-parametric Bayesian) process~\cite{khan2012stick} to address this challenge. Specifically, the probability of the first step (no propagation), i.e., $q({t_{n}}=0)$  is modeled as a break of proportion $\alpha_{0,n}$ (quit at the first-step), while the length of the remainder of the propagation is left for the next break.  Each probability of propagation step can be deterministically computed by the quit probability
$q({t_{n}}=k)=\alpha_{k,n}\prod_{k'=0}^{{k-1}}\left(1-\alpha_{k',n}\right)$ until $K-1$, and the probability of last propagation step is   $q({t_{n}}=K)=\prod_{k'=0}^{{K-1}} (1-\alpha_{k',n})$. Assume the maximum propagation step $K=2$, then the propagation probability  is generated by 2 breaks where $p(t_n=0)=\alpha_{0, n}$, $p(t_n=1)=\alpha_{1,n}\left(1-\alpha_{0,n}\right)$  and the  last propagation step $p(t_n=2)=(1-\alpha_{1,n})\left(1-\alpha_{0,n}\right)$ (not quit until the end). Hence, for different values of $K$, this non-parametric breaking process always satisfies $\sum_{k=0}^{K} q(t_n=k)=1$.
We call this method \textit{Learning to Quit} ({L2Q}).  Compared with {L2S}, \text{L2Q} models the quit probability of each node at each propagation step via the stick-breaking process which naturally induces the sparsity property of the modeling propagation step for each node. The deeper layers are less likely to be sampled. Figure~\ref{fig:framework}(c) shows the architecture of \text{L2Q}.

\subsection{{Learning and Inference}}
\label{Sec:learning}
Maximizing the ELBO in Eq.~(\ref{Eq:ELBO})  is challenging. The lack of labels for test data further exacerbates the difficulty. Thus, in this paper, we propose two algorithms:  the alternate expectation maximization and iterative variational inference algorithms to maximize it.\\
\textbf{Alternate expectation maximization}.
Minimization of the negative ELBO in Eq.~(\ref{Eq:ELBO}) can be  solved by the expectation maximization (EM) algorithm, which iteratively infers $q(t_n)$ at E-step and learns ${\theta}$  at M-step. More specifically, at each iteration $i$, given the current status parameters ${\theta}^{(i)}$, the E-step that maximizes $\mathcal{L}({\theta}^{(i)},q)$ w.r.t q has a closed-form solution the same as Eq.~(\ref{Eq:non-par}):
\begin{linenomath}
\small
\begin{align}
q^{(i+1)}(t_n) =q(t_n | \mathbf{X},\mathbf{A}, y_n)=\frac{p_{\theta^{(i)}}(y_{n} | \mathbf{X},\mathbf{A}, t_n)}{\sum_{t_n=0}^{K} p_{\theta^{(i)}}(y_n |\mathbf{X},\mathbf{A}, t_n)^{}} \label{Eq:non-parametric}.
\end{align}
\end{linenomath}
However, we can not utilize this non-parametric posterior since  the label
$y_{n}$ is not available for the test nodes.  We need  to let the training and testing pipeline be consistent.  Thus, we consider projecting the non-parametric posterior to a  parametric posterior $q_{{\phi}}(t_n|\mathbf{X},\mathbf{A})$ (i.e., L2S or L2Q). We adopt an approximation, which has also been used in the classical wake-sleep algorithm~\cite{hinton1995wake}  by minimizing the forward KL divergence $KL(q^{(i+1)}(t_{n})||q_{{\phi}}(t_n|\mathbf{X},\mathbf{A}))$. Then we can get the following pseudo maximizing  likelihood objective:
\begin{linenomath}
\small
\begin{align}
{\phi}^{(i+1)}=\arg\max_{{\phi}}\mathbb{E}_{q^{(i+1)}(t_n)}\left[\log q_{{\phi}}(t_n|\mathbf{X},\mathbf{A})\right].
\end{align}
\end{linenomath}
Given the parametric posterior $q_{{\phi^{(i+1)}}}(t_n|\mathbf{X},\mathbf{A})$, the M-step optimizes  $\mathcal{L}(\mathbf{\theta}$,
$q_{{\phi^{(i+1)}}}(t_n|\mathbf{X},\mathbf{A}))$ w.r.t ${\theta}$. Since there is no analytical solution for deep neural networks, we
update the model parameters $\theta$ with respect to the ELBO by one step of gradient descent. \\
\textbf{Iterative variational inference.}
Although the alternate expectation maximization algorithm is effective to infer the optimal propagation variable, the alternate EM steps are time-consuming and we need calculating the loss at every layer for each training  node, i.e., the $O(N*(K+1))$ complexity. Thus, we  propose an end-to-end iterative algorithm to minimize  negative ELBO. 
Specifically,  we  introduce the parameterized  posterior $q_{{\phi}}(t_n|\mathbf{X},\mathbf{A})$ (i.e., L2S or L2Q) into Eq.~(\ref{Eq:ELBO}) and directly optimize ELBO using re-parameterization trick~\cite{kingma2013auto}.  We infer the optimal propagation distribution $q_{\phi}(t_n| \mathbf{X},\mathbf{A})$  and learn GNN weights $\theta$ jointly through standard back-propagation from the  ELBO in Eq.~(\ref{Eq:ELBO}). However, the optimal propagation steps  $t$ is  discrete and non-differentiable which makes direct optimization difficult. Therefore, we adopt Gumbel-Softmax Sampling~\cite{jang2016categorical,maddison2016concrete}, which is a simple yet effective way to
substitutes the original non-differentiable sample from a discrete distribution with a differentiable sample from a corresponding Gumbel-Softmax distribution. Specifically, we  minimize  the following negative ELBO in Eq.~(\ref{Eq:ELBO})  with the reparameterization trick~\cite{kingma2013auto}:
\begin{linenomath}
\small
\begin{align}
 \mathcal{L}({\theta},{\phi}) =-\log p_{{\theta}}(\mathbf{y} | GNN (\mathbf{X},\mathbf{A},\hat{t}))+\text{KL}(q_{\phi}(t_n|\mathbf{X},\mathbf{A}) ||p(t_n)), \label{Eq:ELBO2}
\end{align}
\end{linenomath}
where $\hat{t}$ is drawn  from a categorical distribution with the discrete variational  distribution $q_{{\phi}}(t_n|\mathbf{X},\mathbf{A})$ parameterized by $\phi$:
\begin{linenomath}
\small
\begin{align}
\hat{t}_{k}=\frac{\exp (( \log ( q_{{\phi}}(t_n|\mathbf{X},\mathbf{A})[a_k])+g_{k}) / \gamma_g )}{\sum_{k'=0}^{K} \exp ( ( \log(q_{{\phi}}(t_n|\mathbf{X},\mathbf{A})[a_{k'}])+g_{k'}) / \gamma_g )},
\end{align}
\end{linenomath}
where $\left\{g_{k'}\right\}_{k'=0}^{K}$ are i.i.d. samples drawn from the Gumbel (0, 1) distribution, $\gamma_g$ is the softmax
temperature, $\hat{t}_{k}$ is the $k$-th value of sample $\hat{t}$ and  $q_{{\phi}}(t_n|\mathbf{X},\mathbf{A})[a_k]$  indicates the $a_{k}$-th index of $q_{{\phi}}(t_n|\mathbf{X},\mathbf{A})$, i.e., the logit corresponding the  $(a_{k}-1)$-th layer. Clearly, when $\tau>0$, the Gumbel-Softmax distribution is smooth so $\phi$ can be  optimized by standard back-propagation. The  KL term in Eq.~(\ref{Eq:ELBO2}) is respect to two categorical distributions, thus it has a closed form.

\section{Bi-level Variational Inference}
So far, we have proposed the L2P framework and shown how to solve it via variational inference. However, as suggested by previous work~\cite{DBLP:conf/iclr/RongHXH20,feng2020graph},  GNNs  suffer from  over-fitting  due to the scarce label information in the graph domain. In this section, we propose the bilevel variational inference to alleviate the over-fitting issue.
\subsection{The Bi-level Objective}
\label{Main:Bilevel}
For our L2P framework, the introduced inference network for joint learning optimal propagation steps in \text{L2S} and \text{L2Q}  also increases the risk of over-fitting as shown in  experiments (\S~\ref{Exp:bilevel}). To solve the over-fitting issue, we  draw inspiration from  gradient-based meta-learning (learning to learn)~\cite{maclaurin2015gradient,franceschi2018bilevel}. Briefly, the objective of ${\phi}$ is to maximize the ultimate measure of the performance of GNN model $p_{{\theta}}(y| GNN(\mathbf{X}, \mathbf{A},t))$, which is the model performance on a held-out validation set. Formally, this goal can be formulated as the following bi-level optimization problem:
\begin{linenomath}
\small
\begin{align}
\min _{{\phi}}  \mathcal{L}_{\text{val}}\left({\theta}^{*}({\phi}), {\phi}\right) \text { s.t. } {\theta}^{*}({\phi})=\arg\min_{{\theta}} \mathcal{L}_{\text {train}}({\theta}, {\phi}), \label{Eq:bi-level}
\end{align}
\end{linenomath}
where $\mathcal{L}_{\text{val}}\left({\theta}^{*}({\phi}), {\phi}\right)$ and $\mathcal{L}_{\text {train}}({\theta}, {\phi})$ are called upper-level and lower-level objectives on validation  and training sets, respectively. For our L2P framework, the objective is the negative ELBO $\mathcal{L}({\theta},{\phi})$ in Eq.~(\ref{Eq:ELBO}). This bi-level update
is to optimize the propagation strategies of each node so that the GNN model performs best on the validation set. Instead of
using fixed propagation steps, it learns to assign adaptive steps while regularizing  the training of a GNN model to improve  the generalization.   Generally, the bi-level optimization problem has to solve each level to reach a local minimum.  However, calculating the optimal ${\phi}$ requires two nested loops of optimization, i.e., we need to compute the optimal
parameter ${\theta}^{*}({\phi})$ for each ${\phi}$. Thus, in order to control
the computational complexity, we propose an approximate alternating optimization method by updating ${\theta}$ and ${\phi}$ iteratively in the next section. 
\subsection{Bi-level Training Algorithm}
Indeed, in general, there is no closed-form expression of
$\theta$, so it is not possible to directly optimize the upper-level
objective function in Eq.~(\ref{Eq:bi-level}). To tackle this challenge, we propose
an  alternating approximation algorithm to speed up computation in this section.\\
\textbf{Updating  lower level ${\theta}$}. Instead of solving
the lower level problem completely per outer iteration, we fix ${\phi}$ and only take the following  gradient steps over mode parameter ${\theta}$ at the $i$-th iteration:
\begin{linenomath}
\small
\begin{align}
{\theta}^{(i)}={\theta}^{(i-1)}-\eta_{{\theta}} \nabla_{{\theta}} \mathcal{L}_{\operatorname{train}}({\theta}^{(i-1)}, {\phi}^{(i-1)}), \label{Eq:theta}
\end{align}
\end{linenomath}
where $\eta_{\theta}$  is the learning rate for ${\theta}$.\\
\textbf{Updating upper level ${\phi}$}. After receiving the parameter ${\theta}^{(i)}$ (a reasonable approximation of ${\theta}^{(*)}({\phi}$)), we can calculate the upper level objective, and update ${\phi}$ through:
\begin{linenomath}
\small
\begin{align}
{\phi}^{(i)}={\phi}^{(i-1)}-\eta_{{\phi}} \nabla_{{\phi}} \mathcal{L}_{\text{val}}({\theta}^{(i)}, {\phi}^{(i-1)}). \label{Eq:phi}
\end{align}
\end{linenomath}
Note that ${\theta}^{(i)}$ is a function of ${\phi}$ due to Eq.~(\ref{Eq:theta}), we can directly back-propagate the gradient through ${\theta}^{(i)}$ to ${\phi}$.  the $\nabla_{\phi} \mathcal{L}_{\mathrm{val}}(\theta^{(i)}, \phi^{(i-1)})$ can be approximated as (see Appendix~\ref{app:bilevel} for detailed derivations):
\begin{align}
&\nabla_{\phi} \mathcal{L}_{\mathrm{val}}(\theta^{(i)}, \phi^{(i-1)})=\nabla_{\phi} \mathcal{L}_{\mathrm{val}}(\bar{\theta}^{(i)}, \phi^{(i-1)})\label{Eq:gradient} \\
&-\eta_{\theta}\frac{1}{\epsilon} ( \nabla_{\phi} \mathcal{L}_{\operatorname{train}}(\theta^{(i-1)}+\epsilon v, \phi^{(i-1)})- \nabla_{\phi} \mathcal{L}_{\operatorname{train}}(\theta^{(i-1)}, \phi^{(i-1)})),  \nonumber
\end{align}
where $v= \nabla_{\theta} \mathcal{L}_{\mathrm{val}}(\theta^{(i)}, \bar{\phi}^{(i-1)})$, and $\bar{\theta}^{(i)}$ and $\bar{\phi}^{(i-1)}$ means stopping the gradient. This can be easily implemented by maintaining a shadow version of $\theta^{(i-1)}$ at last step,  catching the training loss $\mathcal{L}_{\operatorname{train}}(\theta^{(i-1)}, \phi^{(i-1)})$ and computing the new loss $\mathcal{L}_{\operatorname{train}}(\theta^{(i-1)}+\epsilon v, \phi^{(i-1)})$. When $\eta_{\theta}$ is set to 0 in Eq.~(\ref{Eq:gradient}), the second-order derivative  will disappear, resulting in a first-order approximate. In  experiments in \S~\ref{Exp:bilevel}, we  study the effect of bi-level optimization, and the first- and second-order approximates.

Given the above derivations of gradients, we have the complete L2P algorithm by  alternating the update rules in Eqs.~(\ref{Eq:theta}) and~(\ref{Eq:phi}). The  time complexity mainly depends on the bi-level optimization. For the first-order approximate, the complexity is the same as vanilla GNN methods. 
L2P needs approximately 3 $\times$ training time for the second-order approximate since it needs    extra forward and backward passes of the weight  to compute the bilevel gradient. However, as the experiments in \S~\ref{Exp:bilevel} show, the first-order approximate is sufficient to achieve the best performance.

\section{EXPERIMENT}
In this section, we conduct experiments  to evaluate the effectiveness of the proposed frameworks with comparison to state-of-the-art GNNs. Specifically, we aim to answer the following questions:
\begin{itemize}
\item[\textbf{(RQ 1)}] How effective is the proposed L2P framework for the node classification task on both heterophily and homophily graphs?
\item[\textbf{(RQ 2)}]  Could the proposed L2P  alleviate  over-smoothing?
\item[\textbf{(RQ 3)}] How do the proposed learning algorithms work? Could the bi-level optimization alleviate  the over-fitting issue?
\item[\textbf{(RQ 4)}] Could the proposed  framework adaptively learn propagation strategies  for better understanding the graph structure?
\item[\textbf{(RQ 5)}] Could the proposed L2P framework effectively  the personalized and interpretable propagation strategies for each node?
\end{itemize}

\subsection{Experimental Settings}

\noindent\textbf{Datasets.} We conduct experiments on both homophily and heterophily graphs. For homophily graphs, we adopts three standard citation networks for semi-supervised node classification, i.e., Cora, CiteSeer, and PubMed~\cite{yang2016revisiting}.  Recent studies~\cite{chen2020simple,jin2020node,pei2019geom} show that the  performance of GNNs can significantly drop on heterophily graphs, we also include   heterophily benchmark in our experiments, including Actor, Cornell, Texas, and Wisconsin~\cite{pei2019geom,chen2020simple}.  The  descriptions and statistics of these datasets are provided  in  Appendix~\ref{app:datasets}.

\begin{table}
\centering
\setlength{\abovecaptionskip}{0.1cm}
\setlength{\belowcaptionskip}{-0.5cm}
\caption{Summary of  results on homophily graphs. Note our results can  be easily improved by using a more complex backbone. For example, by using  GCNII* as our backbone, L2S can achieve 85.6 $\pm$ 0.2 on Cora and 80.9 $ \pm$ 0.3 on PubMed.}
\begin{tabular}{lcccc}
       \toprule[1.0pt]
\textbf{Method}      & \textbf{Cora}& \textbf{CiteSeer} & \textbf{PubMed} \\
       \toprule[1.0pt]
GCN        & 81.3 $\pm$ 0.8 & 71.1 $\pm$ 0.7     & 78.8 $\pm$ 0.6     \\
GAT       & 83.0 $\pm$ 0.7 & 72.5 $\pm$ 0.7     & 79.0 $\pm$ 0.3   \\
APPNP    & 83.3 $\pm$ 0.5 & 71.8 $\pm$ 0.5     & 79.7 $\pm$ 0.3   \\
JKNet       & 80.6 $\pm$ 0.5 & 69.6 $\pm$ 0.2     & 77.8 $\pm$ 0.3   \\
JKNet(Drop) & 83.0 $\pm$ 0.3 & 72.2 $\pm$ 0.7     & 78.9 $\pm$ 0.4   \\
Incep(Drop) & 83.0 $\pm$ 0.5 & 72.3 $\pm$ 0.4     & 79.3 $\pm$ 0.3  \\
DAGNN & 84.2 $\pm$ 0.5 & 73.3 $\pm$ 0.6     & 80.3 $\pm$ 0.4  \\
GCNII* & \textbf{85.3 $\pm$ 0.2} & 73.2 $\pm$ 0.8     & 80.3 $\pm$ 0.4  \\
    \toprule[0.7pt]

L2S & 84.9 $\pm$ 0.3 & 74.2 $\pm$ 0.5     & 80.2 $\pm$ 0.5 \\
L2Q & 85.2 $\pm$ 0.5 &\textbf{74.6 $\pm$ 0.4}      & \textbf{80.4 $\pm$ 0.4} \\
       \toprule[1.0pt]
\end{tabular}\label{table:homo}
\end{table}
\begin{table}
\setlength{\abovecaptionskip}{0.1cm}
\setlength{\belowcaptionskip}{-0.5cm}
\caption{Node classification accuracy on heterophily
graphs. }
\centering
\begin{tabular}{lcccccccc}
   \toprule[1.0pt]
\textbf{Method}      & \textbf{Actor} & \textbf{Cornell} & \textbf{Texas} & \textbf{Wisconsin}\\
   \toprule[1.0pt]
GCN         & 26.86   & 52.71  & 52.16 & 45.88 \\
GAT         & 28.45   & 54.32 & 58.38 & 49.41 \\
Geom-GCN-I  & 29.09   & 56.76 & 57.58 & 58.24 \\
Geom-GCN-P  & 31.63      & 60.81 & 67.57 & 64.12 \\
Geom-GCN-S  & 30.30      & 55.68 & 59.73 & 56.67 \\
APPNP       & 32.41     & 73.51 & 65.41 & 69.02 \\
JKNet       & 27.41  & 57.30  & 56.49 & 48.82 \\
JKNet(Drop) &29.21 & 61.08 & 57.30  & 50.59 \\
Incep(Drop) & 30.13& 61.62 & 57.84 & 50.20  \\
GCNII*      & 35.18   & 76.49 & 77.84 & 81.57 \\
   \toprule[0.7pt]
L2S         & 36.58    & 80.54 & 84.12 & 84.31 \\
L2Q         & \textbf{36.97}   &  \textbf{81.08} & \textbf{84.56} &  \textbf{84.70}  \\
\toprule[1.0pt]
\end{tabular}\label{table:full-heter}
\end{table}

\noindent\textbf{Baselines.} To evaluate the  effectiveness of the proposed framework, we consider the following representative and  state-of-the-art GNN models on the semi-supervised node classification task. GCN~\cite{DBLP:conf/iclr/KipfW17}, GAT~\cite{DBLP:conf/iclr/VelickovicCCRLB18}, JK-Net~ \cite{xu2018representation}, APPNP~\cite{DBLP:conf/iclr/KlicperaBG19}, DAGNN \cite{DBLP:conf/nips/Zhou0LZCH20}, IncepGCN  \cite{DBLP:conf/iclr/RongHXH20}, and GCNII* \cite{chen2020simple}. We also compare our proposed methods with GCN(DropEdge), ResGCN(DropEdge), JKNet(DropEdge) and IncepGCN(DropEdge) by utilizing the drop-edge trick~\cite{DBLP:conf/iclr/RongHXH20}.
The details and implementations of baselines  are given in  Appendix~\ref{app:Baselines}.\\
\noindent\textbf{Setup.}
 For our L2P framework, we consider APPNP as our backbone unless otherwise stated, but note that our  framework is  broadly applicable to  more complex GNN backbones~\cite{DBLP:conf/iclr/KipfW17,DBLP:conf/iclr/VelickovicCCRLB18,chen2020simple}. We randomly initialize the model parameters.
  We utilize the first-order approximate for our methods due to its efficiency  and study the effect of second-order approximate separately in \S~\ref{Exp:bilevel}. For each search of hyper-parameter
configuration, we run the experiments with 20 random seeds and
select the best configuration of hyper-parameters based on average accuracy on the validation set. Hyper-parameter settings and the splitting of datasets are given in Appendix~\ref{app:setup}.  
\subsection{{RQ1. Performance Comparison}}
\label{sec:sota}
To answer RQ1, we conduct experiments on both homophily and heterophily graphs with
comparison to state-of-the-art methods.
\noindent\textbf{Performance on homophily graphs.} Table \ref{table:homo} reports the mean classification accuracy with the standard deviation on the test nodes after 10 runs. From Table \ref{table:homo}, we have the following findings: (1) Our L2S and L2Q  improve the performance of the APPAP backbone consistently and significantly in all settings. This is because that our framework has the advantage of adaptively learning personalized strategies via bi-level training. This observation demonstrates our motivation and the effectiveness of our framework. (2) Our L2S and L2Q can achieve comparable performance  with state-of-the-art methods such as DAGNN and GCNII* on Cora and PubMed, and outperform them on CiteSeer. This once again demonstrates the effectiveness of our L2P framework on the  node classification task. (3) In terms of our methods, the L2Q performs  better than L2S, indicting that the simple softmax is not the best parameterization for the variational distribution of the latent propagation variable.  \\
\noindent\textbf{Performance on heterophily graphs.} Besides the previously mentioned baselines, we also compare our methods with three variants of Geom-GCN~\cite{pei2019geom}: Geom-GCN-I, Geom-GCN-P, and Geom-GCN-S. Table~\ref{table:full-heter} reports the results.  (1) We can observe that
\text{L2S} and \text{L2Q}  outperform the APPNP backbone on four heterophily graphs, which indicates our framework can still work well on the heterophily graphs.  (2) L2S and L2Q consistently improve GCNII* by a large margin and achieve new state-of-the-art results on four heterophily graphs. (3) We can  find that the improvement on heterophily graphs is usually larger than that on homophily graphs (Table \ref{table:homo}). This is because the neighborhood information is noisy, aggregating the neighborhood information may result in worse performance for GCNII*. In contrast, our  \text{L2S} and \text{L2Q}  adaptively learn the process of propagation which can avoid utilizing the structure information which maybe not helpful for heterophily graphs. 
\begin{figure}
	\setlength{\abovecaptionskip}{-0.1cm}
  \setlength{\belowcaptionskip}{-0.2cm}
	\centering
	\includegraphics[width=0.7\linewidth]{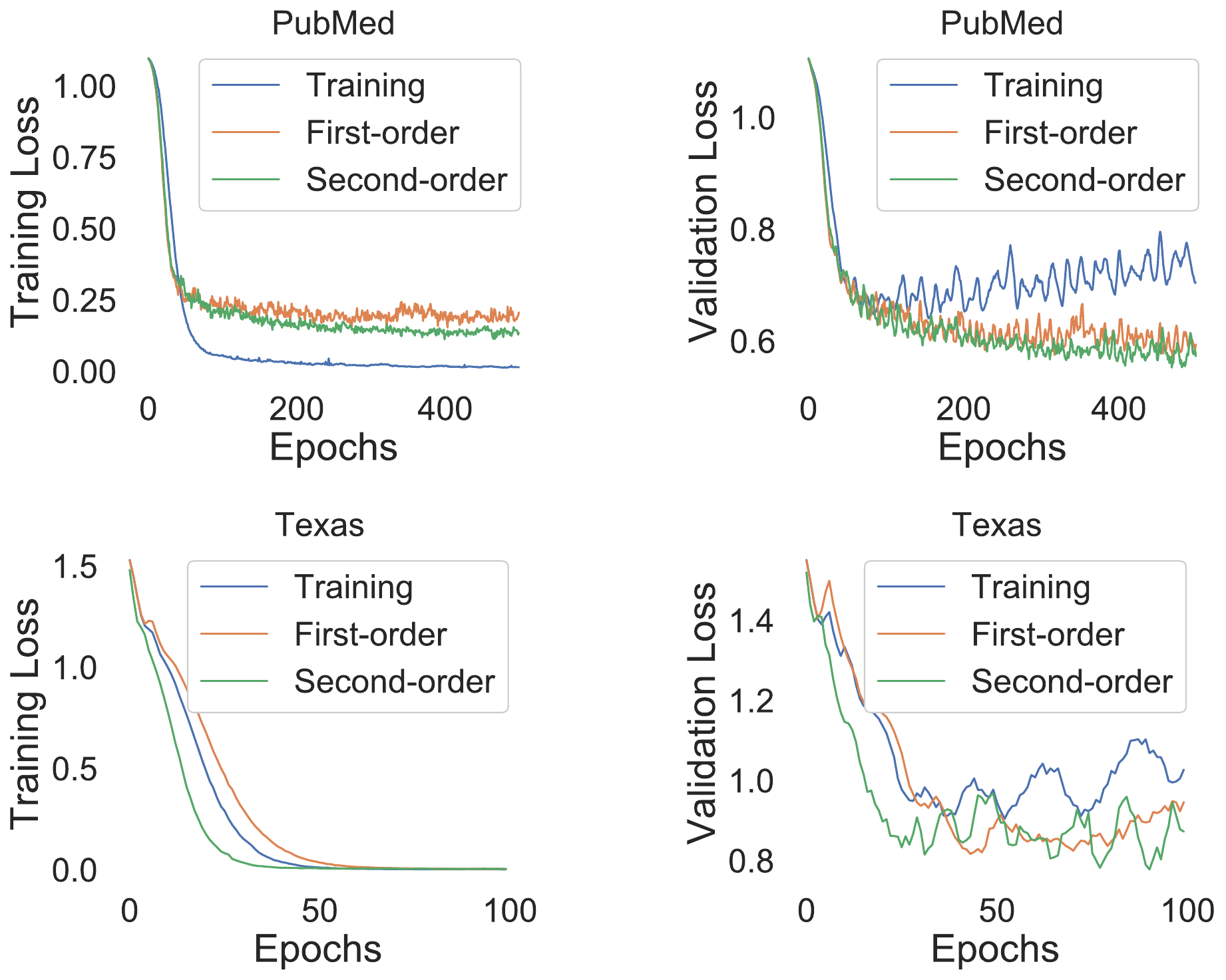}
	\caption{The training  and validation losses of \text{L2Q}.}
	\label{fig:l2qpub}
\end{figure}

\begin{figure}[h]
\setlength{\abovecaptionskip}{-0.2cm}
\setlength{\belowcaptionskip}{-0.2cm}
	\centering
	\includegraphics[width=0.44\linewidth]{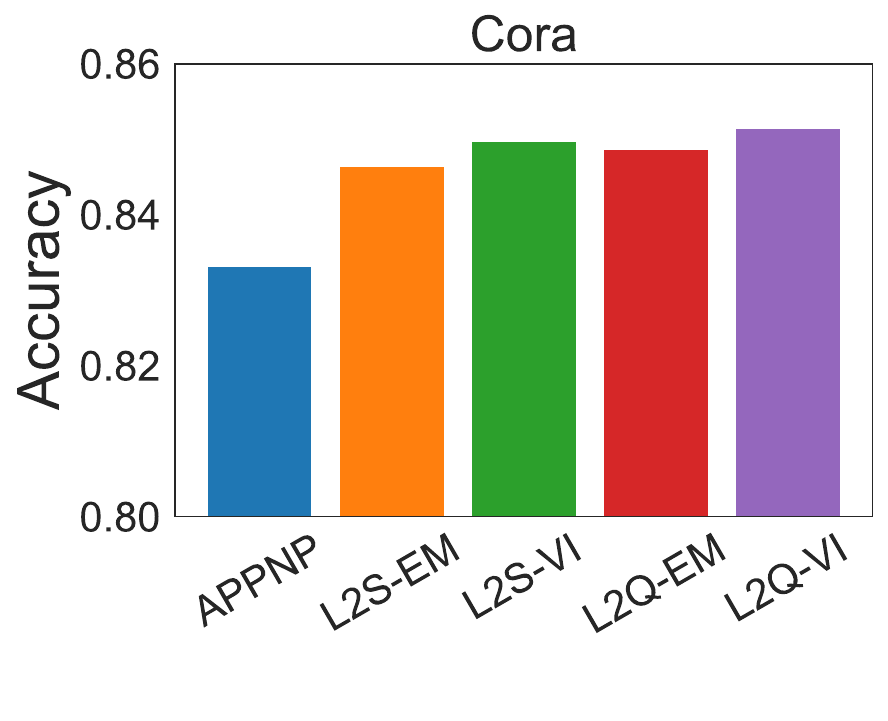}
	\includegraphics[width=0.44\linewidth]{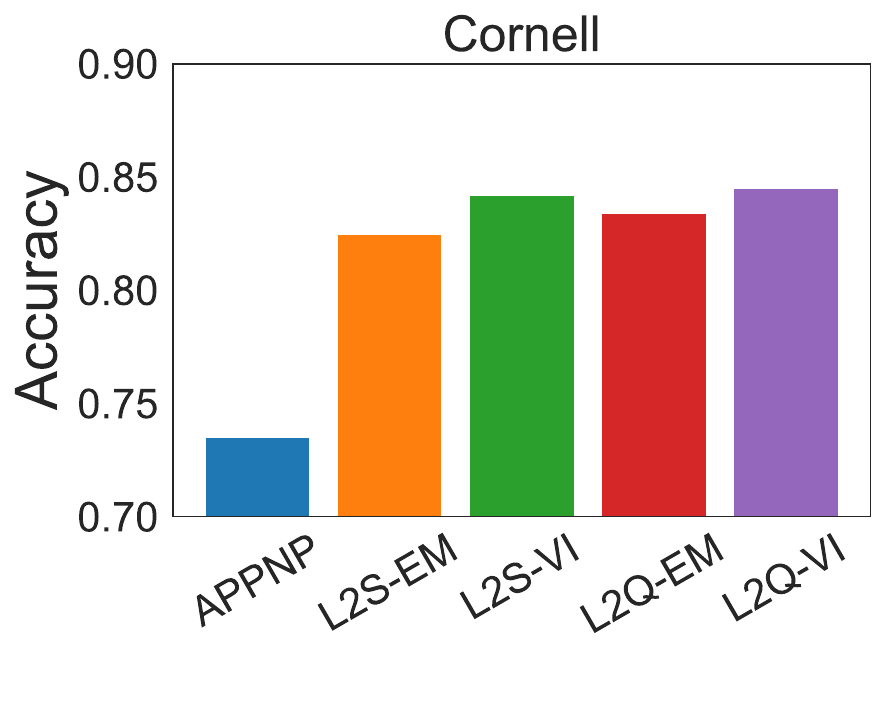}
	\caption{Comparison of different learning algorithms.}
	\label{fig:methods}
\end{figure}

\begin{table}[]
\setlength{\abovecaptionskip}{0.1cm}
\setlength{\belowcaptionskip}{-0.2cm}
\caption{Semi-supervised classification accuracy (\%) on different connections using  GCN as the backbone.}
\centering
\begin{tabular}{lcccccccl}
   \toprule[1.0pt]
\textbf{Dataset}  & \textbf{GCN}    & \textbf{Res} & \textbf{JK} & \textbf{Incep}  & \textbf{L2S} & \textbf{L2Q} \\
   \toprule[1.0pt]
Cora & 81.3 & 78.8 & 81.1  & 81.7    & \textbf{82.6} & 81.5 \\
CiteSeer & 71.1 & 70.5 & 69.8  & 70.2 & 71.3  & \textbf{71.9} \\
PubMed  & 78.8 & 78.6 & 78.1  & 77.9  & 79.4  & \textbf{79.6}  \\
 \toprule[0.7pt]

Actor & 30.3      & 31.3 & 34.2 & 32.4 & 35.0  &\textbf{35.1} \\
Cornell      & 57.0  & 60.2 & 64.6 & 66.5  & 70.2  & \textbf{70.5}\\
Texas         & 59.5  & 65.7 & 66.5 & 75.6  & 80.3  & \textbf{80.5} \\
Wisconsin       & 59.9   & 71.2 & 74.3 & 75.1 & 80.0  & \textbf{80.1} \\
\toprule[1.0pt]
\end{tabular}\label{semi-connect}
\end{table}

\begin{table}
\centering
\setlength{\abovecaptionskip}{-0.0cm}
\setlength{\belowcaptionskip}{-0.4cm}
\caption{Classification accuracy (\%) results with different pre-defined  propagation  steps on Cora.}
\begin{tabular}{lcccccc}
\toprule[1.0pt]
\multirow{2}{*}{\textbf{Method}} & \multicolumn{6}{c}{\textbf{Propagation Steps}}               \\
                        & 2    & 4    & 8    & 16   & 32   & 64   \\
                        \toprule[1.0pt]
GCN                     & 81.1 & 80.4 & 69.5 & 64.9 & 60.3 & 28.7 \\
GCN(Drop)               & \textbf{82.8} & 82.0   & 75.8 & 75.7 & 62.5 & 49.5 \\
JKNet                   & -    & 80.2 & 80.7 & 80.2 & 81.1 & 71.5 \\
JKNet(Drop)             & -    & \textbf{83.3} & 82.6 & 83.0   & 82.5 & 83.2 \\
Incep                   & -    & 77.6 & 76.5 & 81.7 & 81.7 & 80.0   \\
Incep(Drop)             & -    & 82.9 & 82.5 & 83.1 & 83.1 & 83.5 \\
APPNP & 81.8 & 82.0 &  83.2 & 83.8 &  82.6 & 82.0 \\
GCNII*                & 80.2 & 82.3 & 82.8 & 83.5 & 84.9 & \textbf{85.3} \\ \hline
L2S                     & 82.2 & 82.8 & \textbf{84.9} & 84.6 & 84.6 & 84.6 \\
L2Q                     & 82.2 & {83.2} & 84.8 & \textbf{84.8} & \textbf{85.2} & {85.2}
\\ \toprule[1.0pt]
\end{tabular}\label{semilayer}
\end{table}

\noindent \textbf{Performance with the other backbone}.
To further show the effectiveness of our framework, we use the GCN as the backbone and  compare our methods with the following connection designs which toward alleviate the over-smoothing or capture higher-order information: Residual (Res) ~\cite{DBLP:conf/iclr/KipfW17,li2019deepgcns,DBLP:conf/iclr/RongHXH20},  Jumping knowledge (JK) ~\cite{xu2018representation}, and Inception (Incep)~\cite{DBLP:conf/iclr/RongHXH20} connections.  Table~\ref{semi-connect} shows the performance of different connections on homophily and heterophily graphs. From Table~\ref{semi-connect}, we have the following findings: (1) Our L2S and L2Q outperform the baselines, especially in heterophily with GCN backbone, which suggests that our framework is agnostic to backbones and graphs. (3) Although the advanced connections such as Res and JK can alleviate the over-smoothing, they still do not outperform 2-layer GCN on homophily graphs. Our L2S and L2Q are the only two methods that perform better than GCN across all the datasets. These findings show that our L2P framework can effectively adapt to both heterophily and homophily graphs.

\begin{figure}[h]
\setlength{\abovecaptionskip}{-0.05cm}
\setlength{\belowcaptionskip}{-0.1cm}
	\centering
		\includegraphics[width=0.99\linewidth]{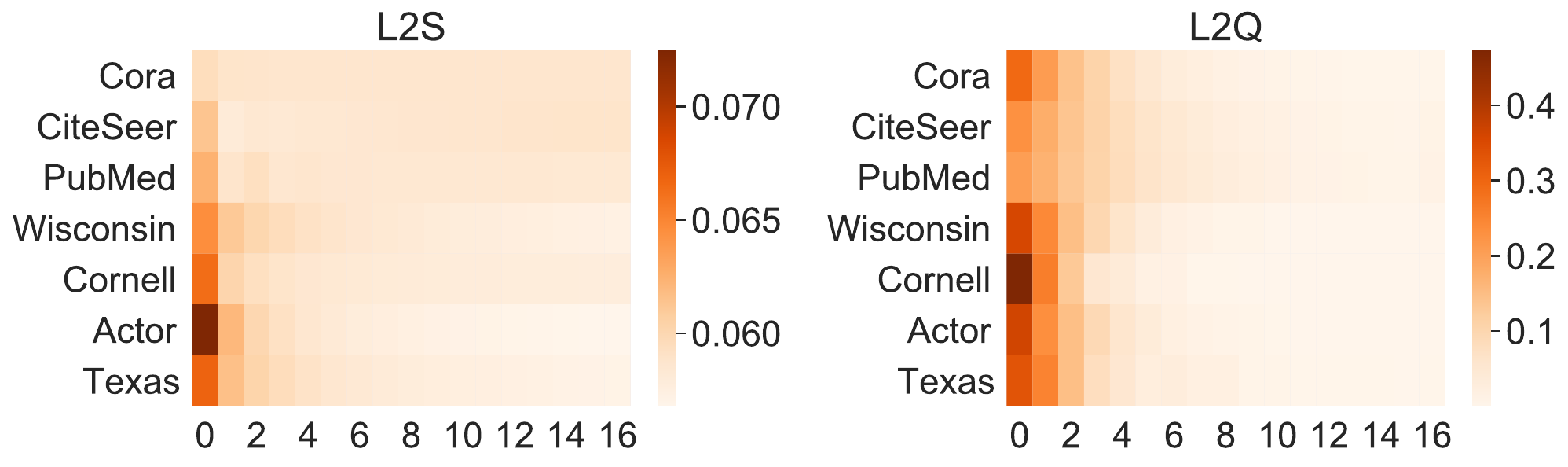}
	\caption{The propagation distributions on different  graphs.}
	\label{fig:propagHeat}
\end{figure}
\begin{figure}
\setlength{\abovecaptionskip}{-0.1cm}
\setlength{\belowcaptionskip}{-0.1cm}
	\centering
	\includegraphics[width=0.99\linewidth]{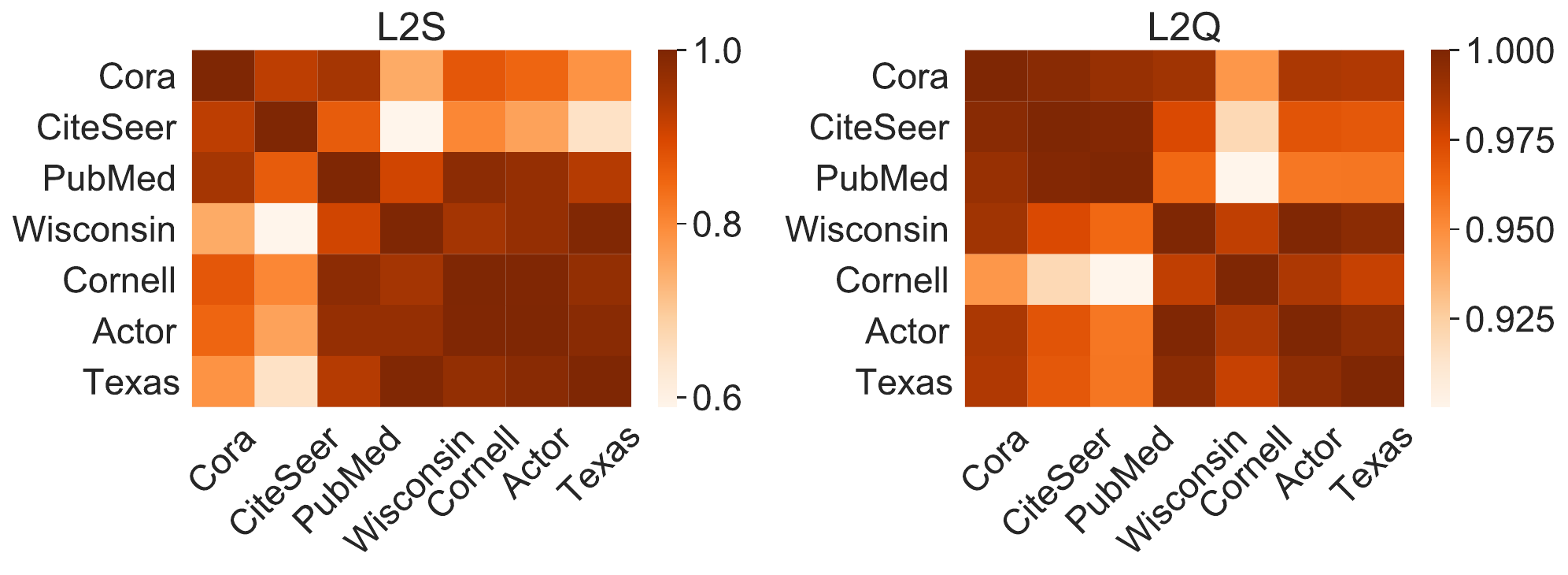}
	\caption{Graph correlation obtained from our learned propagation distributions. Similar graphs are more correlated, such as Cora is
closer to CiteSeer than Texas.}
	\label{fig:corprop}
	
\end{figure}
\begin{figure*}
	\centering
		\includegraphics[width=0.85\linewidth]{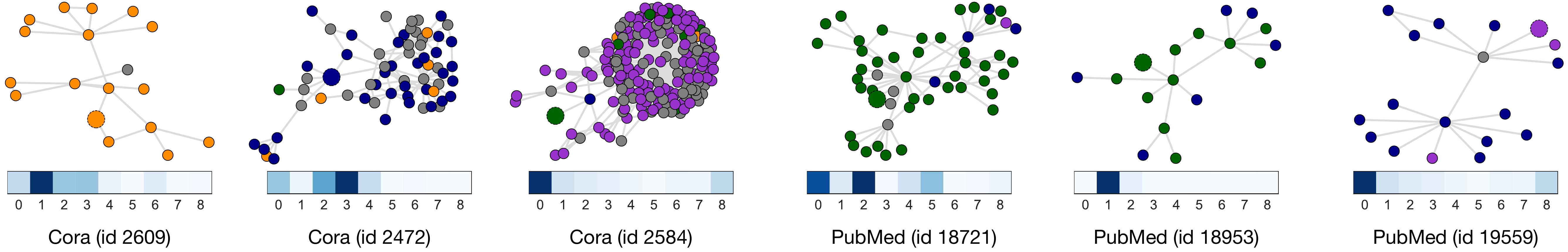}
	\caption{Case studies of the personalized propagation on two homophily datasets. The bigger node in each sub-graph is the test node. The  propagation distributions learned by \text{L2Q} of the test nodes  are visualized with  heatmaps (bottom).}
	\label{fig:Visu1}
\end{figure*}
\begin{figure*}
	\centering
		\includegraphics[width=0.85\linewidth]{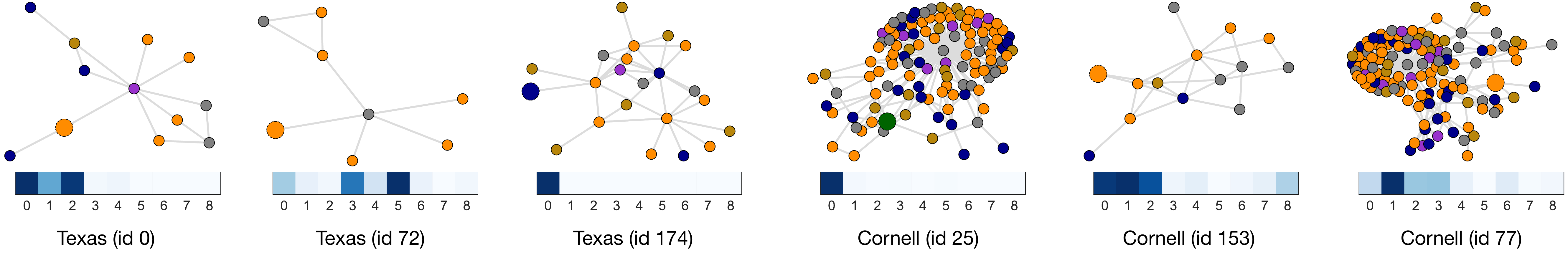}
  \vskip -1em
	\caption{Case studies of the personalized propagation on two heterophily datasets. The bigger node in each sub-graph is the test node. The  propagation distributions learned by \text{L2Q} of the test nodes  are visualized with  heatmaps (bottom).}
	\label{fig:Visu2}
 
\end{figure*}

\subsection{{RQ2.  Over-smoothing}}
To answer \text{RQ2}, we study how the proposed methods perform as the number of layers increases compared to state-of-the-art (deep) GNNs. We vary the number of layers as $\{2,4,8,16,32,64\}$. We only report the performance on Cora  as we have similar observations on other datasets. Table \ref{semilayer} summaries the semi-supervised results for the deep models with various propagation steps. We observe that the performance of proposed methods consistently improves as increasing the number of layers, which indicates the effectiveness of our L2P framework. For all cases, the proposed methods achieve the best accuracy under a depth beyond 2, which again verifies the impact of L2P on formulating graph neural networks. Notably, our methods achieve the best performance as we increase the network depth to 64 and the results of our methods remain stable with stacking many layers. On the other hand, the performance of GCN with DropEdge and JKNet drops rapidly as the number of layers exceeds 32, which represents that they  suffer from  over-smoothing. This phenomenon suggests that with an adaptive and personalized message propagation strategy, L2P can  effectively resolve the over-smoothing problem and achieve  better performance.

\subsection{{RQ3. The Effect of Learning Algorithms}}
\label{Exp:bilevel}
\noindent To answer \text{RQ3}, We first compare the performance of  alternate  expectation maximization (EM) and iterative variational inference (VI). From Figure~\ref{fig:methods}, we can find that our methods with two learning algorithms both achieve better performance  compared to the best results of APPNP, which verifies the effectiveness of our learning algorithm. In general, the iterative VI achieves better performance than the EM algorithm. We then analyze the model loss of stochastic bi-level variational inference with \textit{training}  (we optimize $\phi$ simultaneously with $\theta$ on training data without validation), \textit{first-order} and \textit{second-order} approximates. Figures~\ref{fig:l2qpub} show the learning curves of training loss and validation loss on the Texas and PubMed datasets  of \text{L2Q}.   We can observe that the \textit{training} gets stuck in the overfitting issue attaining low training loss but high validation loss. The gap between training  and validation losses is much smaller for first-order and second-order. This  demonstrates that  the bilevel optimization can significantly improve generalization capability and the first-order approximate is sufficient to prevent the overfitting. 

\subsection{RQ4. Adaptive Propagation Strategies.}
One of the most important properties of our framework is that the learned propagation strategy is interpretable and is different for different types of graphs and nodes. Thus, in this subsection, we investigate if the proposed framework can learn adaptive propagation strategies, which aims to answer \text{RQ4}. 
We visualize the  average propagation distribution (via averaging propagation distributions of all nodes) for seven graphs learned by \text{L2S} and \text{L2Q} with K=16 in Figure~\ref{fig:propagHeat}. The darkness of a step represents the probability that the step is selected for propagation. From Figure~\ref{fig:propagHeat}, we can find that (1) different types of  graphs  exhibit different propagation  distributions although the pre-defined step is 16 for all of them. For instance, the  0-th step probability in heterophily graphs  is much larger than that of homophily graphs. This is because that the feature information in those heterophily graphs is much more important
than the structure information. (2) The propagation distribution learned  by \text{L2Q} is much sparse, and the layers on the tail are less likely to be sampled. In Figure~\ref{fig:corprop},  we also provide the correlation, i.e. the cosine similarity of learned propagation distributions of different graphs.  We clearly observe the  correlations between the same types of graphs are large while the correlation between homophily and heterophily graphs is small, which meets our expectation that similar types of graphs should generally have similar propagation strategies.
%

\subsection{RQ5. Personalized Propagation Strategies}
To evaluate if L2P can learn good personalized and  interpretable  propagation for \text{RQ5}, we  study the propagation strategy for individual nodes. Figures~\ref{fig:Visu1} and~\ref{fig:Visu2} show the case studies of personalized propagation on  homophily and heterophily graphs. In Figures~\ref{fig:Visu1} and \ref{fig:Visu2}, we plot the 3-hop neighborhood of each test node and use different colors to indicate different  labels. We find that a test node with more same class neighbors  tends to propagate few steps.   In contrast, a test node with  fewer same class nodes will probably have more propagation steps to predict truly  its label.  This observation matches our intuition that  different nodes need different propagation strategies, and the prediction of a node will be confused if it has too many propagation steps and thus can not benefit much from message propagation. Additionally, we can find that our framework  successfully identifies the propagation steps that are important for predicting the class of nodes on both homophily and heterophily graphs and has a more interpretable prediction process.

\section{Conclusion}
In this paper, we study the problem of learning the
propagation strategy in GNNs.  We propose \text{learning to propagate} (L2P), a general   framework to address this problem. Specifically,  we introduce the optimal propagation steps as latent variables to help find the maximum-likelihood estimation of the GNN parameters and infer the  optimal propagation step for each node via the VEM. Furthermore, we propose L2S and L2Q, two instances to parameterize the variational propagation distribution and frame the variational inference process as a bi-level optimization problem  to alleviate the
over-fitting problem. Extensive experiments demonstrate that our L2P can achieve  state-of-the-art performance on  seven benchmark datasets and  adaptively capture the personalized and interpretable propagation strategies of different nodes and various graphs.

\begin{acks}
The authors would like to thank the Westlake University and Bright Dream Robotics Joint Institute for the funding support. Suhang Wang is supported by the National Science Foundation under grant number IIS-1909702, IIS1955851, and Army Research Office (ARO) under grant number W911NF-21-1-0198.
\end{acks}


\bibliographystyle{ACM-Reference-Format}
\bibliography{Reference}

\appendix
\section{Appendix}
\subsection{Derivations of the Evidence Lower Bound}
\label{app:ELBO}
\begin{linenomath}
\small
\begin{align}
&\log p_{\theta}\left(\mathbf{y}_{n} | \mathbf{X}, \mathbf{A}\right) = \log \text{$\sum\nolimits_{t_n=0}^{K}$}p_{\theta}\left(\mathbf{y}_{n} | {GNN}(\mathbf{X}, \mathbf{A},t_n)\right)p(t_n) \nonumber \\
&=\mathbb{E}_{q_{}(t_n)}\left[\log p_{\theta}\left(\mathbf{y}_{n} | GNN( \mathbf{X}, \mathbf{A}, t_{n}\right))\right]-\text{KL}(q_{}(t_n) ||p(t_n) )  
 \nonumber \\
&+\text{KL}(q_{}(t_n) ||p(t_n |\mathbf{X}, \mathbf{A}, y_n))\nonumber \\
&\geq \mathcal{L}({\theta}, q)= \mathbb{E}_{q_{}(t_n)}\left[\log p_{\theta}\left(\mathbf{y}_{n} | \mathbf{X}, \mathbf{A}, t_{n}\right)\right]-\text{KL}(q_{}(t_n) ||p(t_n)).\label{Eq:ELBO3}
\end{align}
\end{linenomath}
The inequality holds since the $KL\left(q_{}(t_n) ||p(t_n |\mathbf{X}, \mathbf{A}, y_n)  \right)$ is always no less than zero. The ELBO itself is a lower bound on the log evidence (the log-likelihood), whilst the variational distribution $q(t_n)$ serves as an approximation of the  posterior $p(t_n |\mathbf{X}, \mathbf{A}, y_n)$~\cite{blei2017variational}.
\subsection{Derivations of the Bi-level Gradient}
\label{app:bilevel}
 ${\theta}^{(i)}$ is a function of ${\phi}$ due to Eq.~(\ref{Eq:theta}), we can directly back-propagate the gradient through ${\theta}^{(i)}$ to ${\phi}$. Based on the chain rule, the gradient $ \nabla_{{\phi}} \mathcal{L}_{\text{val}}({\theta}^{(i)}, {\phi}^{(i-1)})$ can be approximated as follows:
\begin{linenomath}\small
\begin{align}
&\nabla_{\phi} \mathcal{L}_{\mathrm{val}}(\theta^{(i)}, \phi^{(i-1)}) \label{App:bilevel} \\
&=\nabla_{\phi} \mathcal{L}_{\mathrm{val}}(\bar{\theta}^{(i)}, \phi^{(i-1)})+\nabla_{\phi} \mathcal{L}_{\mathrm{val}}(\theta^{(i)}, \bar{\phi}^{(i-1)}) \nonumber\\
&=\nabla_{\phi} \mathcal{L}_{\mathrm{val}}(\bar{\theta}^{(i)}, \phi^{(i-1)})+
\nabla_{\theta^{(i)}} \mathcal{L}_{\mathrm{val}}(\theta^{(i)}, \bar{\phi}^{(i-1)})\nabla_{\phi}\theta^{(i)}(\phi)\nonumber \\
&=\nabla_{\phi} \mathcal{L}_{\mathrm{val}}(\bar{\theta}^{(i)}, \phi^{(i-1)})+ \nonumber \\
&\nabla_{\theta^{(i)}} \mathcal{L}_{\mathrm{val}}(\theta^{(i)}, \bar{\phi}^{(i-1)})\nabla_{\phi}({\theta}^{(i-1)}
-\eta_{{\theta}} \nabla_{{\theta}} \mathcal{L}_{\operatorname{train}}({\theta}^{(i-1)}, {\phi}^{(i-1)}))= \nonumber \\
&\nabla_{\phi} \mathcal{L}_{\mathrm{val}}(\bar{\theta}^{(i)}, \phi^{(i-1)}) -\eta_{\theta} \nabla_{\theta} \mathcal{L}_{\mathrm{val}}(\theta^{(i)}, \bar{\phi}^{(i-1)}) \nabla_{\phi} \nabla_{\theta} \mathcal{L}_{\operatorname{train}}(\theta^{(i-1)}, \phi^{(i-1)}).  \nonumber 
\end{align}
\end{linenomath}
In the last line, we make a Markov assumption that $\nabla_{\phi} \theta^{i-1} \approx 0$, assuming that at iteration step $i$, given $\theta_{i-1}$ we do not care about how the values of $\phi$ from previous steps led to $\theta_{i-1}$. This assumption can decrease the   computation cost, and it has already shown empirical success in prior works on the bi-level optimization~\cite{baydin2018online,baydin2018automatic}. For the second-order term $ \nabla_{\theta} \mathcal{L}_{\mathrm{val}}(\theta^{(i)}, \bar{\phi}^{(i-1)})\nabla_{\phi}  \nabla_{\theta} \mathcal{L}_{\operatorname{train}}(\theta^{(i-1)}, \phi^{(i-1)})$, we further propose an efficient  approximation of  it  by utilizing  the first-order Taylor expansion of $\nabla_{\theta} \nabla_{\phi} \mathcal{L}_{\operatorname{train}}$
$(\theta^{(i-1)}, \phi^{(i-1)})$. Specifically, for any vector $v\in \mathbb{R}^{|\theta|}$, with small $\epsilon>0$, we have:
\begin{align}
&v^{\top} \cdot \nabla_{\theta} \nabla_{\phi} \mathcal{L}_{\operatorname{train}}(\theta^{(i-1)}, \phi^{(i-1)}) \label{App:Taylor}\\
&\approx \frac{1}{\epsilon} ( \nabla_{\phi} \mathcal{L}_{\operatorname{train}}(\theta^{(i-1)}+\epsilon v, \phi^{(i-1)})- \nabla_{\phi} \mathcal{L}_{\operatorname{train}}(\theta^{(i-1)}, \phi^{(i-1)}) ). \nonumber
\end{align}
Given this,  $\nabla_{\phi} \mathcal{L}_{\mathrm{val}}(\theta^{(i)}, \phi^{(i-1)})$ in Eq.~(\ref{App:bilevel}) can be approximated  as:
\begin{align}
&\nabla_{\phi} \mathcal{L}_{\mathrm{val}}(\bar{\theta}^{(i)}, \phi^{(i-1)})-\eta_{\theta}\nabla_{\theta} \mathcal{L}_{\mathrm{val}}(\theta^{(i)}, \phi^{(i-1)})\cdot \nonumber \\
&\nabla_{\phi}  \nabla_{\theta} \mathcal{L}_{\operatorname{train}}(\theta^{(i-1)}, \phi^{(i-1)})=\nabla_{\phi} \mathcal{L}_{\mathrm{val}}(\bar{\theta}^{(i)}, \phi^{(i-1)})\label{App:second2} \\
&-\eta_{\theta}\frac{1}{\epsilon} ( \nabla_{\phi} \mathcal{L}_{\operatorname{train}}(\theta^{(i-1)}+\epsilon v, \phi^{(i-1)})- \nabla_{\phi} \mathcal{L}_{\operatorname{train}}(\theta^{(i-1)}, \phi^{(i-1)})),  \nonumber
\end{align}
where $v= \nabla_{{\theta}} \mathcal{L}_{\mathrm{val}}(\theta^{(i)}, \bar{\phi}^{(i-1)})$.

\subsection{Datasets Description and Statistics}
\label{app:datasets}
In our experiments, we use the following real-world datasets. The statistics of datasets are given in Table~\ref{dataStat}.\\
\begin{table}
\centering
\caption{The statistics of datasets.}\label{dataStat}
\vspace{-0.2cm}
\begin{tabular}{ccccccc}
\toprule[1.0pt]
\textbf{Dataset}   & \textbf{Classes} & \textbf{Nodes}  & \textbf{Edges}  & \textbf{Features} \\
\toprule[1.0pt]
Cora      & 7       & 2,708  & 5,429  & 1,433    \\
Citeseer  & 6       & 3,327  & 4,732  & 3,703    \\
PubMed    & 3       & 19,717 & 44,338 & 500      \\
Actor     & 5       & 7,600  & 26,659 & 932      \\
Texas     & 5       & 183    & 309    & 1,703    \\
Cornell   & 5       & 183    & 295    & 1,703    \\
Wisconsin & 5       & 251    & 499    & 1,703  \\ \toprule[1.0pt]
\end{tabular}
\vspace{-0.5cm}
\end{table}
\textbf{Cora, PubMed and CiteSeer} are citation and homophily graphs, which are among the most widely used benchmarks for semi-supervised node classification~\cite{yang2016revisiting,DBLP:conf/iclr/KipfW17,DBLP:conf/iclr/VelickovicCCRLB18}. In these citation datasets, nodes are documents, and edges are citations. Each node is
assigned a class label based on the research field. These datasets use a bag of words representation as the feature vector for each node.\\
\textbf{Actor} is a heterophily graph representing actor co-occurrence in Wiki pages~\cite{pei2019geom} based on
the film-director-actor-writer network. \\
\textbf{Texas, Wisconsin and Cornell} are heterophily graphs representing links between web pages of the corresponding universities, originally collected by the CMU WebKB project. We use the preprocessed datasets in~\cite{pei2019geom}. These datasets are web networks, where nodes and edges represent web pages and hyperlinks, respectively.
\subsection{Baselines and Implementations}
\label{app:Baselines}
\textbf{GCN}: GCN~\cite{DBLP:conf/iclr/KipfW17} is a widely used graph convolutional model.\\
\textbf{GAT}: GAT~\cite{DBLP:conf/iclr/VelickovicCCRLB18} utilizes the attention mechanism and assigns different weights to different neighborhoods in the propagation step. \\
 \textbf{JK-Net}: JK-Net~ \cite{xu2018representation} utilizes dense connections to leverage different neighbor ranges for better representations of nodes.\\
  \textbf{APPNP}: APPNP  \cite{DBLP:conf/iclr/KlicperaBG19} adds the original node feature to the representation learned by each layer, which can well preserve the personalized information of nodes so as to alleviate over-smoothing.\\
 \textbf{DAGNN}: DAGNN \cite{DBLP:conf/nips/Zhou0LZCH20} proposes adaptive weighting  to integrate representations from different aggregation steps into the last layer. \\
 \textbf{IncepGCN}: IncepGCN  \cite{DBLP:conf/iclr/RongHXH20} utilizes  inception backbones with graph convolution layers to capture the information from different hops. \\
\textbf{GCNII}: GCNII \cite{chen2020simple} improves GCN by adding residual connection and identity mapping. We compare our methods with GCNII* which is a variant of GCNII employing two weight matrices, since it can generally achieve better performance than GCNII as shown in \cite{chen2020simple}. \\
\textbf{Implementations.} For all baselines, we used the official implementation publicly released by the authors on Github.\\
\textbf{Hardware.} We ran our experiments on  GeForce RTX 2080 Ti (11G).
\subsection{Experimental setup}
\label{app:setup}
\textbf{Dataset splitting}. For homophily graphs (Cora, PubMed, and CiteSeer), we follow the widely used semi-supervised setting in~\cite{yang2016revisiting,DBLP:conf/iclr/KipfW17,DBLP:conf/iclr/VelickovicCCRLB18} and apply the standard fixed training/validation/testing split  with 20 nodes per class for training, 500 nodes for validation and 1,000 nodes for testing.  For heterophily graphs, we use the feature vectors,
class labels, and 10 random splits (48\%/32\%/20\% of nodes
per class for train/validation/test) from~\cite{zhu2020beyond,pei2019geom}.\\
\textbf{Parameter setting}.
We randomly initialize the parameters. For our methods, the hyper-parameter search spaces are as follows:  dropout (0.2, 0.4, 0.6), learning rate (0.001, 0.005, 0.01), hidden layer size (64, 128), L2 weight-decay (5e-4, 1e-4, 5e-6, 1e-6). For all methods. the propagation steps K is tuned from $(2, 4, \dots, 32, 64)$. For each search of hyper-parameter configuration, we run the experiments with 20 random seeds and select the best configuration of hyper-parameters based on average accuracy on the validation set.

\end{document}